\documentclass[letterpaper]{article} 
\usepackage{aaai25}  
\usepackage{times}  
\usepackage{helvet}  
\usepackage{courier}  
\usepackage[hyphens]{url}  
\usepackage{graphicx} 
\usepackage{float}
\usepackage{natbib}  
\usepackage{caption} 
\usepackage{algorithm}
\usepackage{algorithmic}
\usepackage{xcolor}
\usepackage{placeins}
\usepackage{placeins}

\usepackage{newfloat}
\usepackage{listings}
\DeclareCaptionStyle{ruled}{labelfont=normalfont,labelsep=colon,strut=off} 
\floatstyle{ruled}
\newfloat{listing}{tb}{lst}{}
\floatname{listing}{Listing}
\nocopyright
\title{Tiny-YOLOSAM: Fast Hybrid Image Segmentation}
\author{
    Kenneth Xu\textsuperscript{\rm *},
    Songhan Wu\textsuperscript{\rm *}
}
\affiliations{
    University of Michigan\\
    \textsuperscript{\rm *}Equal contribution.\\
    \{kennethx, wuumaa\}@umich.edu
}

\begin{document}

\maketitle
\begin{abstract}
The Segment Anything Model (SAM) enables promptable, high-quality segmentation but is often too computationally expensive for latency-critical settings. TinySAM is a lightweight, distilled SAM variant that preserves strong zero-shot mask quality, yet its ``segment-everything'' mode still requires hundreds of prompts and remains slow in practice. We first replicate TinySAM on COCO val2017 using official checkpoints, matching the reported AP within 0.03\%, establishing a reliable experimental baseline. Building on this, we propose Tiny-YOLOSAM, a fast hybrid pipeline that uses a recent YOLO detector (YOLOv12) to generate box prompts for TinySAM on salient foreground objects, and supplements uncovered regions with sparse point prompts sampled only where YOLO-guided masks provide no coverage. On COCO val2017, the hybrid system substantially improves class-agnostic coverage (AR from 16.4\% to 77.1\%, mIoU from 19.2\% to 67.8\%) while reducing end-to-end runtime from 49.20s/image to 10.39s/image (4.7$\times$) on an Apple M1 Pro CPU. These results suggest detector-guided prompting combined with targeted sparse sampling as an effective alternative to dense ``segment-everything'' prompting for practical full-scene segmentation.
\end{abstract}

\section{Introduction}
Foundation vision models have made image segmentation far more general and user-friendly. In particular, the Segment Anything Model (SAM) produces high-quality masks from simple prompts such as points or boxes, enabling flexible zero-shot segmentation across diverse images. However, SAM's strong performance comes with substantial inference cost, which limits its use in real-time and resource-constrained deployments (e.g., robotics, mobile devices, and interactive applications). TinySAM addresses this issue by distilling SAM into a compact model that retains competitive zero-shot segmentation quality. Despite being significantly smaller and faster than SAM, TinySAM still faces a practical bottleneck in its ``segment-everything'' setting: achieving full-scene coverage requires dense prompting and repeated decoder calls, resulting in high end-to-end latency.

To reduce redundant prompts while preserving mask quality, recent work has explored detector-guided prompting: using a fast object detector to propose regions of interest, then applying a promptable segmentation model only where needed. This idea is attractive because it replaces blind dense sampling with object-aware proposals, often yielding large speedups. However, purely detector-guided segmentation inherits the detector's recall limitations: objects outside the detector's label space or low-confidence regions may be missed, and background regions (e.g., sky, terrain, buildings) can remain unsegmented. Motivated by this trade-off, we propose a hybrid design that (i) uses detector boxes to efficiently segment salient foreground instances, and (ii) recovers missed/background regions via lightweight sparse prompting \emph{only in uncovered areas}. Compared to hierarchical dense prompting, this targeted uncovered-region sampling reduces wasteful decoder queries, while mitigating the coverage gap of detector-only approaches such as MobileSAMv2.

In this work, we present \textbf{Tiny-YOLOSAM}, a fast hybrid segmentation pipeline that couples a recent YOLO detector (YOLOv12) with TinySAM. The pipeline operates in three stages: (1) YOLO generates bounding box proposals for prominent foreground objects; (2) each box is used as a prompt to TinySAM to produce high-resolution instance masks; (3) a binary coverage mask is constructed from these foreground masks, and sparse point prompts are sampled \emph{only} from uncovered regions to prompt TinySAM for background and miscellaneous regions, after which overlapping masks are merged via IoU-based post-processing.

Our implementation and reproducibility materials are available in our public repository.\footnote{\url{https://github.com/Kenneth-Xu11566/tiny-yolosam}} This manuscript is a revised and streamlined version of a course project report for \textit{EECS 498: Machine Learning Research Experience} at the University of Michigan (\textit{Fall 2025}), with updated presentation and consolidated experimental reporting.\footnote{\url{https://github.com/498ers/Tiny-YOLOSAM_Paper/releases/tag/course-submission-v1}}

Our contributions are as follows:
\begin{itemize}
    \item \textbf{Reproducible TinySAM replication.} We validate the TinySAM evaluation pipeline on COCO val2017 using official checkpoints, and report matched AP results together with hardware and runtime measurement details for reproducibility.
    \item \textbf{A detector-prompted segmentation baseline with analysis.} We implement a YOLO$\rightarrow$TinySAM box-prompted baseline, and analyze how detector recall and prompt count affect runtime, while segmentation quality is further measured through mIoU and AP.
    \item \textbf{Uncovered-region sparse prompting for full-scene coverage.} We propose an uncovered-region sampling strategy that places sparse point prompts only where detector-guided masks provide no coverage, and introduce a class-agnostic evaluation protocol (AR/mIoU) to quantify coverage beyond detector label spaces.
\end{itemize}

Empirically, on COCO val2017, our hybrid system substantially improves class-agnostic scene coverage, increasing AR from 16.4\% to 77.1\% and mIoU from 19.2\% to 67.8\%, while reducing end-to-end runtime from 49.20s/image to 10.39s/image (4.7$\times$) on an Apple M1 Pro CPU. For detector-based evaluation on COCO instance categories, our box-prompted pipeline achieves 40.7\% AP (with 17.6\% / 44.8\% / 50.0\% for small/medium/large objects), compared to 46.7\% AP for a stronger proposal generator (ViTDet$\rightarrow$TinySAM). This gap is primarily driven by detector recall: missing small or infrequent objects reduces the number of valid box prompts, and those missed regions may also be partially suppressed by the foreground coverage mask before sparse prompting. Overall, the results indicate a clear speed--coverage advantage over dense ``segment-everything'' prompting, while revealing a precision ceiling set by the detector in category-aware metrics. These observations motivate future improvements such as fine-tuning the detector for higher recall, using adaptive uncovered-region sampling, and introducing category-aware post-processing to reduce detector-induced inadequate evaluation.

\subsection{Related Work}
\begin{enumerate}
    \item \textbf{MobileSAMv2.}
    A substantial fraction of SAM-style ``segment-everything'' runtime arises from dense prompting and repeated inference over many candidate prompts, followed by filtering and merging. MobileSAMv2 mitigates this cost by replacing grid-search prompting with object-aware prompt sampling: it first performs object discovery (e.g., via detector-like proposals) and then feeds only valid prompts to the decoder, substantially reducing redundant decoding and reported to cut decoder time by at least $16\times$. A practical limitation is that its coverage becomes tied to the quality and recall of the detector's object discovery, which can miss small or rare instances. Our approach follows the same high-level motivation (avoid dense prompting) but couples detector-guided box prompts with additional sparse prompting over uncovered regions to recover background and missed areas while maintaining efficiency. \cite{mobilesam2023}

    \item \textbf{FastSAM.}
    FastSAM reframes ``segment anything'' as a two-stage pipeline: a real-time instance segmentation model (implemented with a YOLO-style segmentation branch) first generates masks for all instances, and a lightweight prompt-based selection stage then chooses masks of interest. This design avoids the heavy transformer image encoder used by SAM and enables real-time usage in many settings. However, the final mask quality and coverage are fundamentally bounded by the underlying segmentation detector, and generalization beyond its training label space can be limited. In contrast, our pipeline uses a detector primarily for efficient proposals, while delegating mask refinement to a promptable segmentation model (TinySAM) and supplementing non-detector regions via sparse prompting. \cite{fastsam2023}

    \item \textbf{YOLO-style Instance Segmentation (YOLO-seg).}
    YOLO-based segmentation models extend single-stage detectors with a mask head to predict instance masks directly in one forward pass, offering a strong speed--accuracy trade-off for closed-set instance segmentation on COCO-like categories. Compared to promptable segmentation models, YOLO-seg methods typically focus on category-bounded segmentation and may struggle to represent background regions or out-of-distribution concepts without retraining. Our work uses YOLO-style detection as an efficient proposal mechanism, but relies on a promptable decoder for higher-fidelity mask refinement and extends coverage via sparse prompts on uncovered regions. \cite{yolov8seg}

    \item \textbf{EfficientViT-SAM.}
    EfficientViT-SAM accelerates SAM by retaining SAM's prompt encoder and mask decoder while replacing the heavy ViT image encoder with an EfficientViT backbone, trained via distillation and end-to-end fine-tuning. This line of work is largely orthogonal to ours: EfficientViT-SAM reduces the \emph{per-call} cost of promptable segmentation, whereas our method reduces the \emph{number of calls/prompts} required for full-scene coverage through detector-guided prompting and uncovered-region sparse sampling. Combining a faster SAM-style backbone with our prompting strategy could further improve end-to-end latency. \cite{efficientvitsam2024}
\end{enumerate}

\section{Background}

\subsection{Image Segmentation and SAM}
Image segmentation partitions images into meaningful regions.
The Segment Anything Model (SAM) \cite{kirillov2023sam} introduced prompt-guided universal segmentation,
where users specify regions via points, boxes, or masks (subsequent works have extended to texts and audio).
SAM uses a Vision Transformer (ViT) encoder and a lightweight mask decoder to produce high-quality masks with
zero-shot generalization. However, SAM's large encoder (632M parameters) results in slow inference, limiting
real-time applications.

\subsection{TinySAM}
TinySAM \cite{tinysam2023} addresses SAM's computational cost through three techniques:
(1) \textit{knowledge distillation} from SAM to a smaller TinyViT student,
(2) \textit{quantization} to 8-bit integers, reducing model size by $4\times$,
and (3) \textit{hierarchical segmenting everything}, which samples a coarse grid ($8\times8$ points)
followed by adaptive refinement in uncertain regions.
Despite achieving 42.3\% AP on COCO with significant speedup over SAM, TinySAM still requires dense prompting
(hundreds of points per image); its hierarchical segmenting everything mode would take 31--44 seconds per image
in our experiments.

\subsection{YOLO for Object Detection}
The YOLO (You Only Look Once) family performs single-stage object detection, directly predicting bounding boxes
and classes in one forward pass. YOLOv12 \cite{yolov12} introduces Transformer backbones
with \textit{Area Attention} and \textit{Flash Attention} for efficient memory access. YOLOv12s achieves 47.6\% mAP on COCO with
only 3.1ms latency. While YOLO excels at detecting prominent objects, it only outputs coarse bounding boxes and
may miss objects outside its training categories.

\subsection{Evaluation Metrics}
We use COCO 2017 val split for evaluation. \cite{coco2014} Key metrics include: \textit{Average Precision (AP)} at IoU 0.50:0.95 for instance segmentation, further subdivided by object sizes: small, medium, and large; \textit{Average Recall (AR)} measuring coverage; \textit{Mean IoU (mIoU)} for mask quality. Other than precision metrics provided by COCO, we use \textit{inference time (seconds/image)} and \textit{calls of decoder} to measure the models' speed.

\section{Methodology, Results, and Discussion}

\subsection{TinySAM Replication}
We first replicate TinySAM's baseline performance to verify the reliability of our experimental environment. Using official pretrained weights (both non-quantized and quantized models), we evaluate on the full COCO val2017 dataset (5000 images). Hardware: Apple M1 Pro Silicon.

\begin{table}[ht]
\centering
\caption{TinySAM Replication Results on COCO val2017}
\label{tab:tinysam_replication}
\small
\setlength{\tabcolsep}{1.5pt}
\begin{tabular}{lccc}
\hline
\textbf{Model} & \textbf{AP@0.50:0.95} & \textbf{Paper Claim} & \textbf{Error} \\
\hline
TinySAM (Non-Quantized) & 42.27\% & 42.3\% & 0.03\% \\
TinySAM (Quantized W8A8) & 41.37\% & 41.4\% & 0.03\% \\
\hline
\end{tabular}
\end{table}

Error within 0.03\%, we successfully confirmed the correctness of our experimental setup and evaluation pipeline.

\subsection{Hierarchical Segment-Everything Baseline}
TinySAM adopts a two-round hierarchical ``segment-everything'' strategy: it first runs mask generation with a coarse point grid (e.g., \texttt{points\_per\_side}=8, i.e., $8\times 8=64$ points; $1/4$ of the default per-side density), filters high-confidence masks to mark covered regions, and then re-samples points at the original density (default \texttt{points\_per\_side}=32, i.e., $32\times 32=1024$ points over the full image) only on the remaining regions. The two rounds of masks are finally merged with post-processing.

Key performance characteristics include a processing time of 2.89 seconds per image, 13.1 TinySAM calls per image, and coverage metrics of AR=16.4\% and mIoU=19.2\%.

\textit{Notes on Hardware and Batching: Hierarchical baseline performance varies significantly across platforms: 49.20s on M1 Pro (CPU) vs. 2.89s on RTX 4070 (GPU with batching). We report both values for transparency, using the M1 Pro measurements as our primary comparison baseline to maintain consistency with hybrid system evaluation.}

While this method achieves full-image segmentation, it suffers from two critical issues: (1) extremely expensive computational overhead, unsuitable for practical applications; (2) poor coverage and segmentation quality, missing many ground truth objects.

Full-scene segmentation is slow because models like SAM and TinySAM must evaluate hundreds of prompts across the whole image. MobileSAMv2 \cite{mobilesam2023} showed that using a detector to guide these prompts can make segmentation dramatically more efficient. Building on this idea, we investigate whether pairing YOLO with TinySAM can create a lightweight, fast, and accurate “segment everything’’ pipeline.

\subsection{Naive Extension: YOLO$\rightarrow$TinySAM}
Our first extension explores a straightforward integration of YOLOv12 and TinySAM, leveraging the strengths of each model to accelerate instance segmentation while retaining high mask quality. The central idea is to replace TinySAM’s expensive hierarchical “segment everything” procedure with objectness-aware box prompts supplied directly by YOLO. This enables TinySAM to focus exclusively on the most salient regions of an image, dramatically reducing the number of prompts and decoder calls required for mask generation.

The naive pipeline consists of three primary stages. First, YOLOv12s-Turbo processes the input image and outputs a set of bounding boxes, class labels, and confidence scores. Second, each YOLO detection box is converted into a box prompt and passed to TinySAM’s mask decoder, which produces a high-resolution segmentation mask for the detected object. Since TinySAM excels at refining object outlines within a spatially constrained region, this step provides detailed, pixel-accurate masks without requiring dense sampling of the entire image. Finally, duplicate masks are removed using IoU-based non-maximum suppression, and the resulting instance masks form the complete segmentation output.

\begin{figure*}[t]
    \centering
    \includegraphics[width=\textwidth]{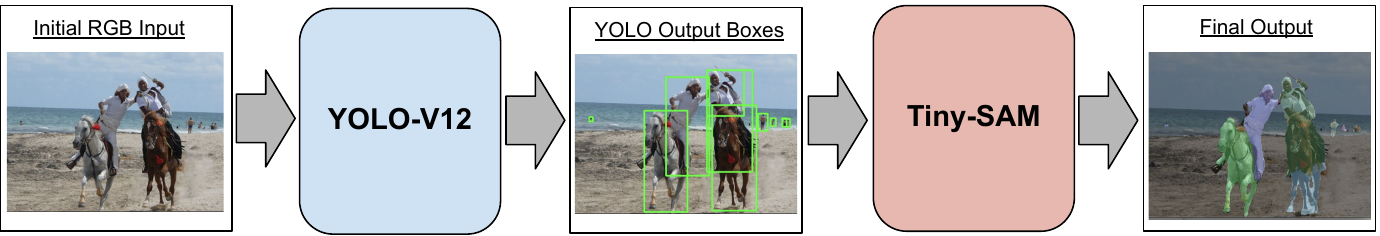}
    \caption{Pipeline for the naive YOLO$\rightarrow$TinySAM extension. 
    YOLOv12 first detects objects and outputs bounding boxes. 
    Each bounding box is then used as a box prompt for TinySAM, 
    which generates instance segmentation masks for detected objects.}
    \label{fig:naive_yolo_tinysam_pipeline}
\end{figure*}

Detector-guided prompting offers several advantages over TinySAM’s hierarchical segment-everything baseline. Most notably, it provides a 7.6× speedup (0.38s vs. 2.89s per image) and dramatically improved mask quality (mIoU 0.93 vs. 0.19) for those masks that are comparable, while requiring only 8.1 decoder calls compared to the hierarchical method’s 13.1.

\begin{table}[ht]
\centering
\caption{Hierarchical v.s. YOLO$\rightarrow$TinySAM Comparison}
\label{tab:yolo_vs_hierarchical}
\small
\setlength{\tabcolsep}{1.5pt}
\begin{tabular}{lccc}
\hline
\textbf{System} & \textbf{mIoU} & \textbf{Time/img} & \textbf{Calls/img} \\
\hline
Hierarchical & 0.19 & 2.89 & 13.1 \\
\textbf{YOLO$\rightarrow$TinySAM} & \textbf{0.93} & \textbf{0.38} & \textbf{8.1} \\
\hline
\end{tabular}
\end{table}

Unlike the core TinySAM paper, which relies on coarse-to-fine refinement across the entire image, our method uses YOLO to reduce redundant sampling by targeting only semantically meaningful regions. Compared to MobileSAMv2, which uses YOLOv8 box prompts to accelerate SAM, our approach leverages YOLOv12’s stronger detection performance and pairs it with TinySAM’s compact decoder to achieve a more favorable efficiency–accuracy tradeoff for real-time usage.

These results demonstrate that detector-guided prompting is a practical and highly efficient alternative to hierarchical sampling, establishing a  YOLO to TinySAM pipeline as a strong baseline for real-time, instance-aware segmentation. However, this approach is fundamentally limited by YOLO’s recall: it segments only COCO object categories (\textasciitilde{}80 species) and completely misses background regions such as the sky, terrain, or surrounding buildings. This motivates our hybrid extension, which incorporates sparse point prompts to recover background coverage and achieve full-scene segmentation.

\vspace{0.25em}

\vspace{-\baselineskip}  
\subsection{Hybrid Extension: YOLO + Sparse Points}
After evaluating the limitations of the YOLO→TinySAM pipeline, particularly its inability to segment background or non-COCO categories, we introduce a hybrid approach designed to recover full-scene segmentation while retaining detector-guided efficiency. The core idea is to let YOLO handle salient foreground objects, while delegating the remaining unsegmented regions of the image to TinySAM through sparse point prompts.

Our hybrid method begins with YOLOv12 detecting prominent foreground objects, which are segmented by TinySAM using box prompts. We then construct a binary coverage mask that marks all pixels already explained by these foreground masks. Instead of sampling dense grids, we place a lightweight 16×16 sparse grid only over the uncovered regions, producing roughly 80–150 point prompts depending on scene complexity. These points prompt TinySAM to generate background and miscellaneous object masks that YOLO cannot detect. Finally, we merge the foreground and background masks by removing duplicate or heavily overlapping masks using an IoU threshold, ensuring that each region of the image is assigned a single, clean segmentation. This pipeline preserves the speed and semantic precision of the detector while recovering the comprehensive coverage characteristic of segment-everything systems, showing inspirations inherited from the original hierarchical segment-everything design \textit{(Appendix A)}.

\begin{figure*}[t]
    \centering
    \includegraphics[width=\textwidth]{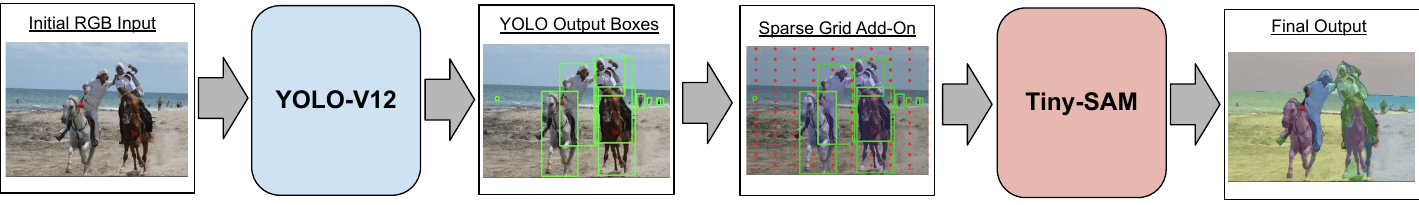}
    \caption{Pipeline for the hybrid YOLO + sparse points extension. 
    YOLOv12 first detects salient foreground objects, which are segmented by TinySAM 
    using box prompts. A coverage mask is then constructed to mark regions already 
    segmented. Sparse points are sampled only in the remaining uncovered regions 
    and passed to TinySAM as point prompts to recover background and small objects. 
    All masks are finally merged using IoU-based post-processing to form full-scene 
    segmentation.}
    \label{fig:hybrid_yolo_tinysam_pipeline}
\end{figure*}

Quantitatively, the hybrid approach substantially improves coverage compared to both Naive(YOLO-only) and TinySAM’s hierarchical baseline.

Before we delve into the evaluation, it is necessary to make it clear why we divide our extension's evaluation metrics into two parts: Detector-based and Class-agnostic. Because \texttt{category\_id} is a necessary parameter for COCO evaluation, original TinySAM manually added ViTDet in the pre-processing stage to provide them (so original TinySAM's images come with box prompts and their \texttt{category\_id} provided by ViTDet). In our extension, for foreground objects with YOLO bounding boxes, we have \texttt{category\_id} inherently, making COCO metrics feasible. Thus, this ``Detector-based" part is evaluated by \textit{AP}. However, for background masks generated by sparse points, we don't have the \texttt{category\_id}, we choose to use Class-agnostic metrics, such as \textit{AR} and \textit{mIoU}.

On 500 COCO val2017 images, for the Class-agnostic objects, average recall increases from 16.4\% to 77.1\% (4.7× improvement), and mIoU rises from 19.2\% to 67.8\% (3.5× improvement). Although the hybrid system is slower than the YOLO-only pipeline due to additional TinySAM calls (about 174 calls per image), it still achieves a 4.7× speedup over the hierarchical baseline (10.39s vs. 49.20s). In Detector-based AP metrics, the hybrid system also demonstrates efficiency gains, achieving a 4.3× speedup over ViTDet→TinySAM (0.38s vs. 1.63s) with fewer decoder calls (8.1 vs. 17.6). However, YOLO→TinySAM's \textit{AP} remains 6 percentage points below the ViTDet→TinySAM baseline, primarily because YOLOv12 misses 38.1\% of small or infrequent objects. Because these objects have already been included (yet not detected) in the `YOLO regions', they will also not be exposed to later background sparse grid detection.

\begin{table}[ht]
\centering
\caption{Detector-based Evaluation (COCO Instance Segmentation Metrics)}
\label{tab:detector_based_eval}
\small
\begin{tabular}{lcccc}
\hline
\textbf{System} & \textbf{AP} & \textbf{AP\_s} & \textbf{AP\_m} & \textbf{AP\_l} \\
\hline
ViTDet$\rightarrow$TinySAM & 46.7\% & 28.9\% & 49.7\% & 59.4\% \\
\textbf{Hybrid (boxes only)} & \textbf{40.7\%} & \textbf{17.6\%} & \textbf{44.8\%} & \textbf{50.0\%} \\
\hline
\end{tabular}
\end{table}

\begin{table}[ht]
\centering
\caption{Detector-based Evaluation (Speed Metrics)}
\label{tab:detector_speed}
\small
\begin{tabular}{lcc}
\hline
\textbf{System} & \textbf{Time/img} & \textbf{Calls/img} \\
\hline
ViTDet$\rightarrow$TinySAM & 1.63 & 17.6 \\
\textbf{Hybrid (boxes only)} & \textbf{0.38} & \textbf{8.1} \\
\hline
\end{tabular}
\end{table}

\begin{table}[ht]
\centering
\caption{Class-agnostic Evaluation (Coverage-focused Metrics)}
\label{tab:class_agnostic_eval}
\small
\setlength{\tabcolsep}{3pt}
\begin{tabular}{lccccc}
\hline
\textbf{System} & \textbf{AR} & \textbf{mIoU} & \textbf{Time/img} & \textbf{Calls/img} \\
\hline
Hierarchical & 16.4\% & 19.2\% & 49.20 & 13.1 \\
\textbf{Hybrid} & \textbf{77.1\%} & \textbf{67.8\%} & \textbf{10.39} & \textbf{174.2} \\
\hline
\end{tabular}
\end{table}

\FloatBarrier
\begin{center}
\textit{Notes: Visualized Graph of Table 3, 4, 5 in Appendix B}
\end{center}

On the other hand, Qualitative results illustrate these gains clearly. In a beach horse-riding scene, the hierarchical baseline produces numerous fragmented masks, while the YOLO-only pipeline captures only the horses and people, ignoring most of the environment. The hybrid method, in contrast, segments both the foreground subjects and large background areas such as the ocean and shoreline, achieving substantially higher recall rate and visual completeness.

\begin{figure}[ht]
    \centering
    \includegraphics[width=\linewidth]{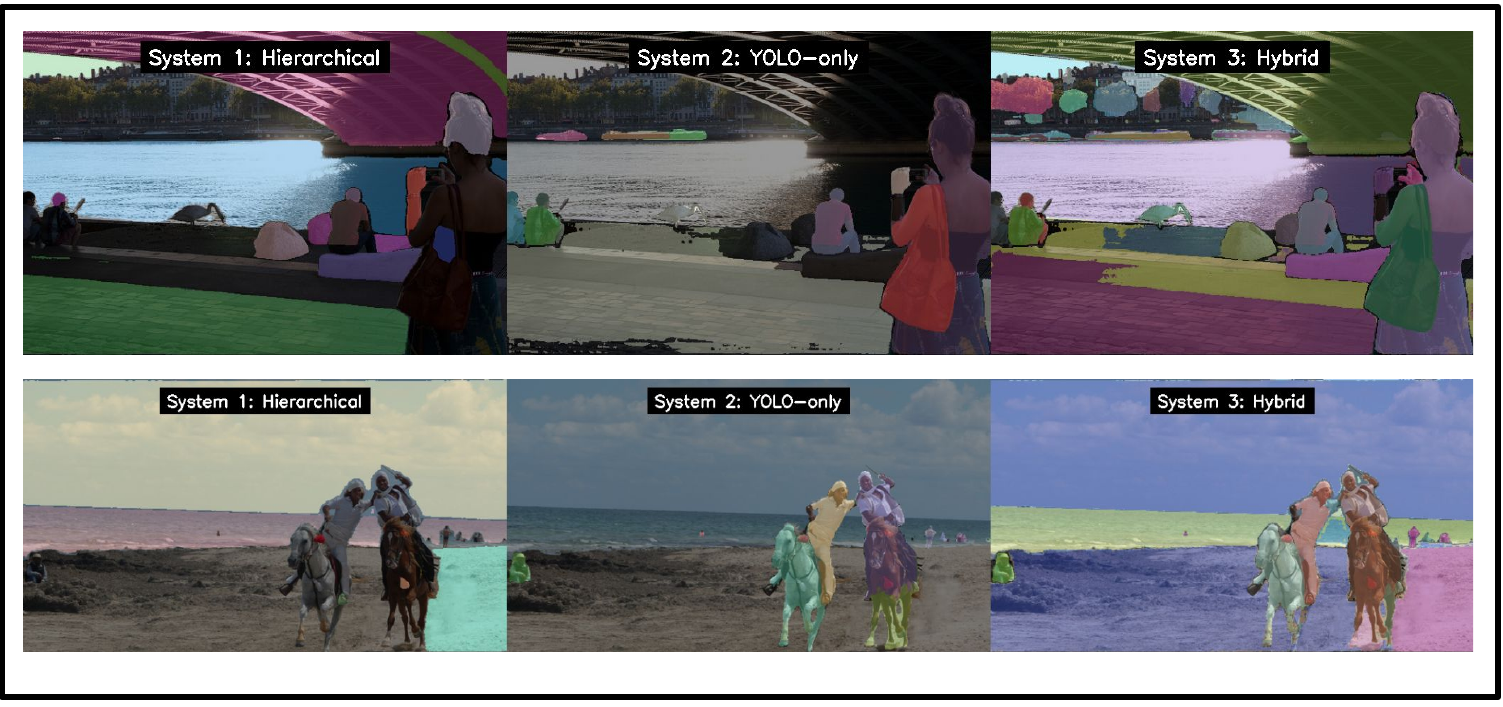}
    \caption{
    Visual comparison of segmentation quality across the three systems evaluated in our 
    study. \textbf{Left: Hierarchical TinySAM} produces coarse and fragmented masks due to 
    dense but low-confidence sampling. \textbf{Middle: YOLO-only} captures prominent 
    foreground objects but fails to segment background regions or smaller structures. 
    \textbf{Right: Hybrid (YOLO + Sparse Points)} recovers both foreground and background 
    content, producing substantially more complete scene coverage. These examples highlight 
    the main advantage of our extension: combining fast detector-guided prompts with sparse, 
    targeted sampling yields both higher coverage and more coherent full-scene segmentation.
    }
    \label{fig:tiny_yolosam_comparison}
\end{figure}

\subsection{Discussion}
Our results highlight both the strengths and limitations of pairing YOLOv12 with TinySAM for fast, full-scene segmentation. On the positive side, the hybrid approach benefits from several algorithmic and engineering advantages. Algorithmically, using YOLO’s single-pass detector to guide segmentation avoids the blind dense sampling required in TinySAM’s hierarchical mode. This significantly reduces redundant computation and eliminates the multi-round refinement steps that slow down the original pipeline. The complementary use of sparse point prompts for background regions further ensures that both foreground and non-COCO objects are covered, enabling a more complete “segment everything’’ capability.

From an engineering standpoint, the modular structure of the hybrid pipeline makes it easy to optimize and debug. Each component, from YOLO for detection to TinySAM for segmentation, can be independently improved or swapped out. Both models rely on mature, well-documented open-source implementations. This contributes to the strong reproducibility of our experiments and lowers the barrier for others to extend or deploy the system.

However, the hybrid system also has notable limitations that stem primarily from the detector. As mentioned, for Detector-based objects, the YOLO$\rightarrow$TinySAM pipeline achieves 40.7\% AP, which is lower than the 46.7\% achieved by ViTDet$\rightarrow$TinySAM. On the full COCO dataset of 5000 images, this issue becomes more pronounced, with AP dropping to 10.7\%. Additionally, direct comparisons with the hierarchical baseline can be misleading. For example, in images containing only large objects, YOLO tends to generate holistic masks (e.g., an entire person), while the hierarchical method often produces fine-grained part-level masks such as clothing or accessories. These reflect different notions of segmentation granularity, making a direct comparison difficult.

Overall, the hybrid model provides substantial gains in speed and coverage, but its precision remains bounded by detector performance. These findings suggest that future improvements should focus on boosting YOLO’s recall, especially for small or uncommon categories, or incorporating additional cues—such as adaptive sampling or semantic post-processing—to mitigate detector omissions.

\subsubsection{Future Work}
\begin{enumerate}
    \item \textbf{Fine-tune YOLO}: Fine-tune YOLOv12 on COCO dataset to improve detection rates for low-frequency categories and small objects, expected to boost AP to 45\%+;
    \item \textbf{Adaptive Grid Density}: Dynamically adjust sparse point sampling based on YOLO coverage density;
    \item \textbf{Category-aware Post-processing}: Use CLIP to unify segmentation granularity for fairer comparison;
    \item \textbf{Video Extension}: Leverage temporal information to improve segmentation consistency.
\end{enumerate}

\section{Conclusion}
We presented \textbf{Tiny-YOLOSAM}, a hybrid full-scene segmentation pipeline that combines fast detector proposals with promptable mask refinement. We first introduced a detector-guided prompting strategy in which YOLOv12 box proposals serve as efficient prompts to TinySAM for high-quality foreground instance masks. To address the coverage limitations of detector-only prompting, we further proposed an \emph{uncovered-region} sparse prompting mechanism that samples point prompts only where the detector-guided masks provide no coverage, enabling more complete scene segmentation without reverting to exhaustive dense prompting.

Empirically, on COCO val2017, our hybrid system improves class-agnostic coverage substantially (AR 16.4\%$\rightarrow$77.1\%, mIoU 19.2\%$\rightarrow$67.8\%) while reducing end-to-end runtime from 49.20s/image to 10.39s/image (4.7$\times$) on an Apple M1 Pro CPU. For category-aware COCO instance segmentation metrics, the box-prompted pipeline reaches 40.7\% AP, which trails a stronger proposal generator (ViTDet$\rightarrow$TinySAM at 46.7\% AP), highlighting a precision ceiling imposed by detector recall, especially for small or infrequent objects. Overall, Tiny-YOLOSAM demonstrates that combining detector-guided box prompts with targeted uncovered-region prompting can deliver a favorable speed--coverage trade-off for practical full-scene segmentation.

Future work will focus on improving proposal recall (e.g., detector fine-tuning and multi-scale inference), designing adaptive uncovered-region sampling to reduce unnecessary prompts, and incorporating lightweight category inference or post-processing to better align class-agnostic masks with category-aware evaluation when needed.
\subsection{Societal Impact}
Tiny-YOLOSAM reduces dense ``segment-everything'' prompting by focusing computation on detector proposals and uncovered regions, improving the practicality of full-scene segmentation on resource-constrained hardware. At the same time, faster segmentation may also lower the barrier for large-scale video analytics and surveillance. However, our pipeline inherits detector limitations (e.g., missed small/rare objects) and may degrade under domain shifts; we recommend validating on target data and adopting privacy-conscious deployment practices.

\bibliography{aaai25}

\section{APPENDIX}
\subsection{Appendix A: Sampling Strategy Comparison}
\begin{figure}[ht]
\centering
\includegraphics[width=\linewidth]{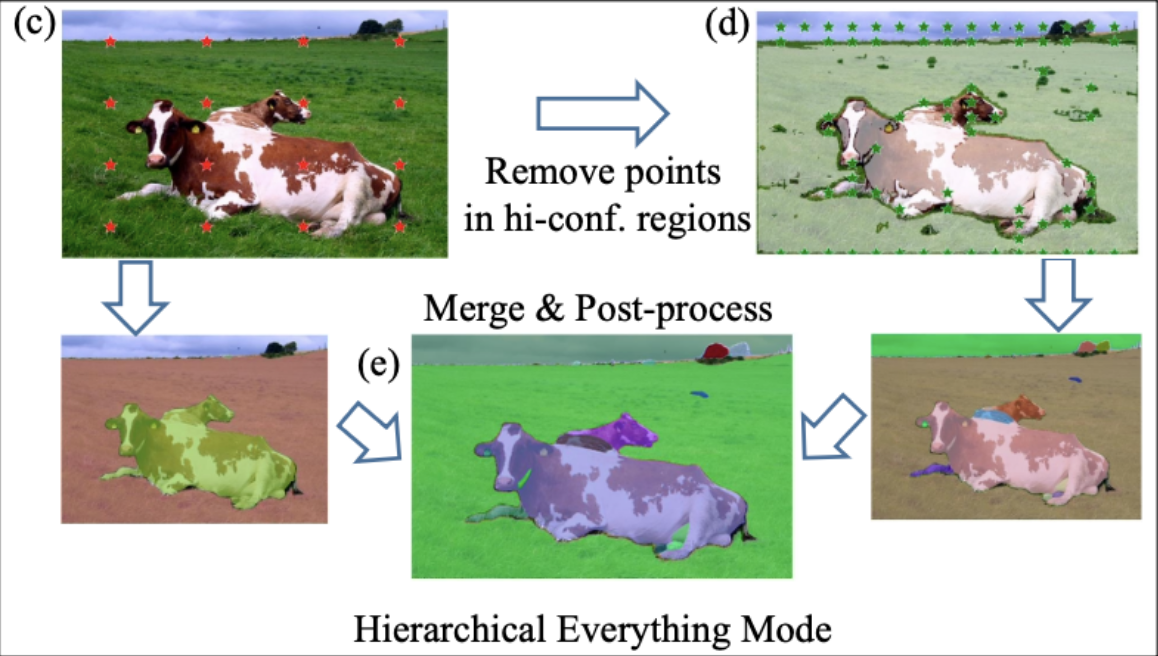}

\vspace{0.5em}

\includegraphics[width=\linewidth]{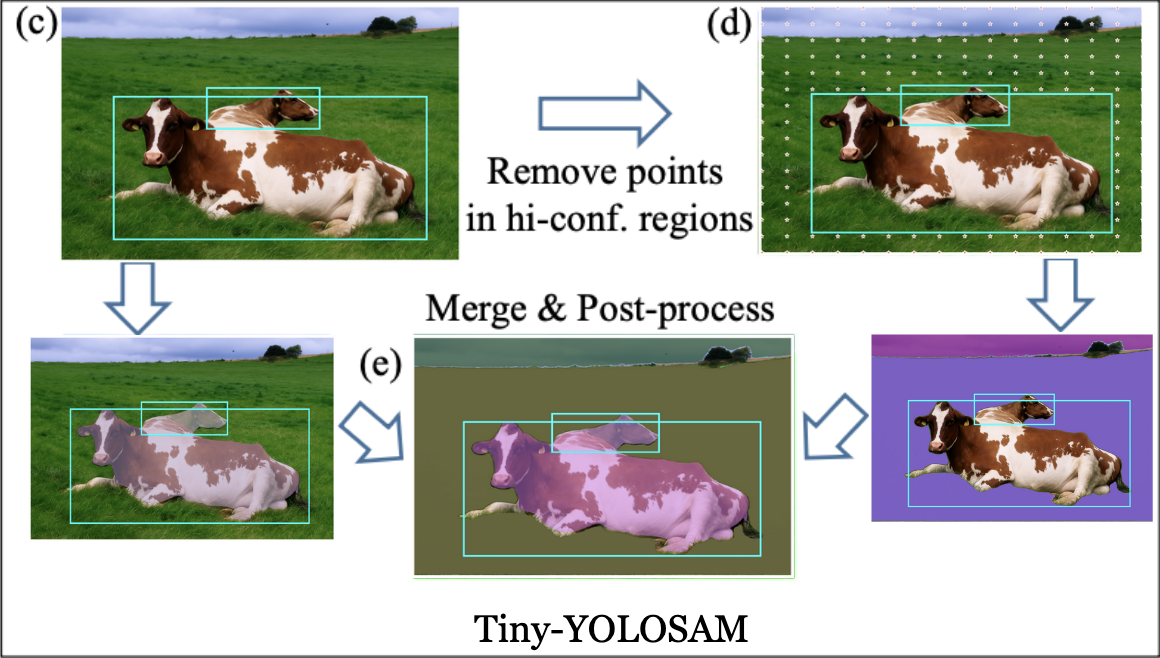}
\caption{Comparison of hierarchical segment-everything prompting and Tiny-YOLOSAM prompting.}
\label{fig:prompting_comparison}
\end{figure}

Both methods follow a similar workflow: (c) initial prompts are provided, (d) high-confidence regions are identified and removed from subsequent sampling, and (e) remaining areas are processed to generate final masks. The key difference lies in the prompting strategy: Hierarchical mode uses a dense 8×8 point grid covering the entire image and performs adaptive resampling in low-confidence areas, resulting in fragmented masks. In contrast, Tiny-YOLOSAM leverages YOLO-detected bounding boxes to segment foreground objects and applies sparse grid sampling only to uncovered background regions, producing more complete and coherent scene segmentation.

\subsection{Appendix B: Quantitative Performance Comparison}
\begin{figure}[htbp]
\centering
\includegraphics[width=2\linewidth]{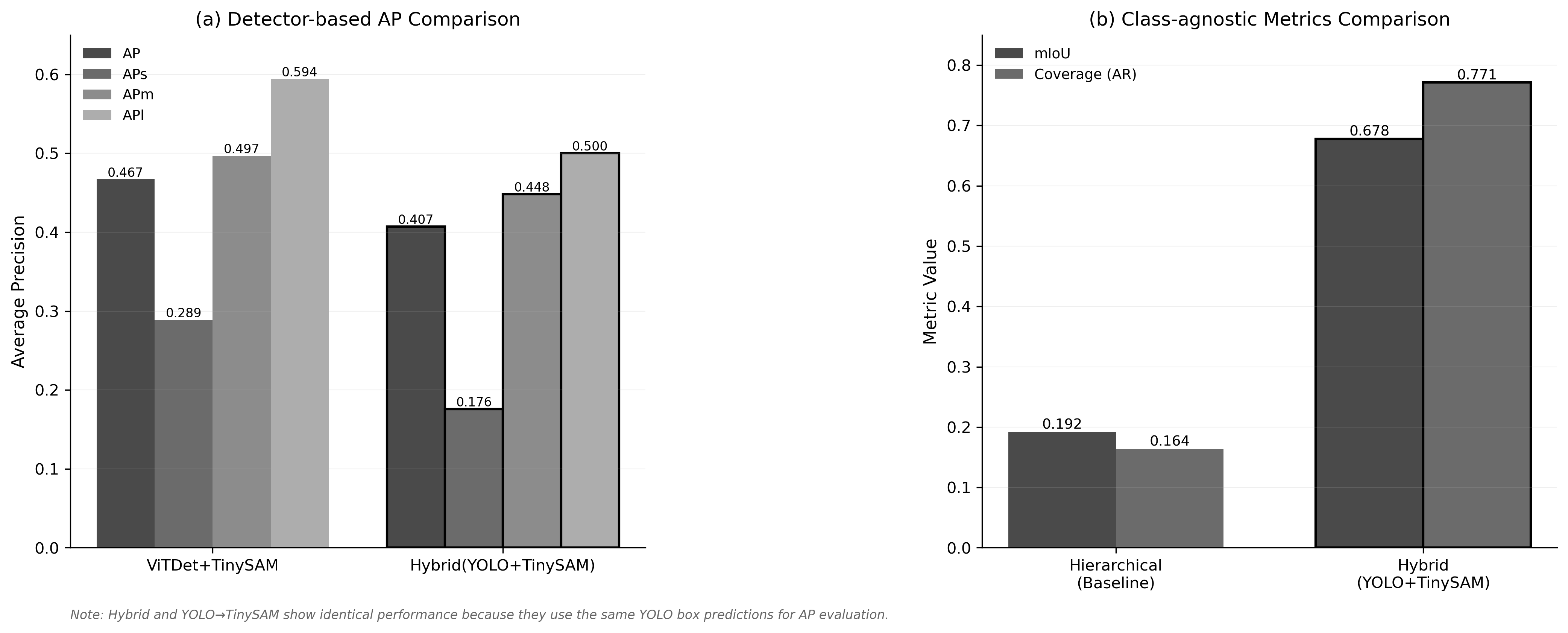}
\caption{Performance comparison.}
\label{fig:metrics_comparison}
\end{figure}

\begin{figure}[htbp]
\centering
\includegraphics[width=2\linewidth]{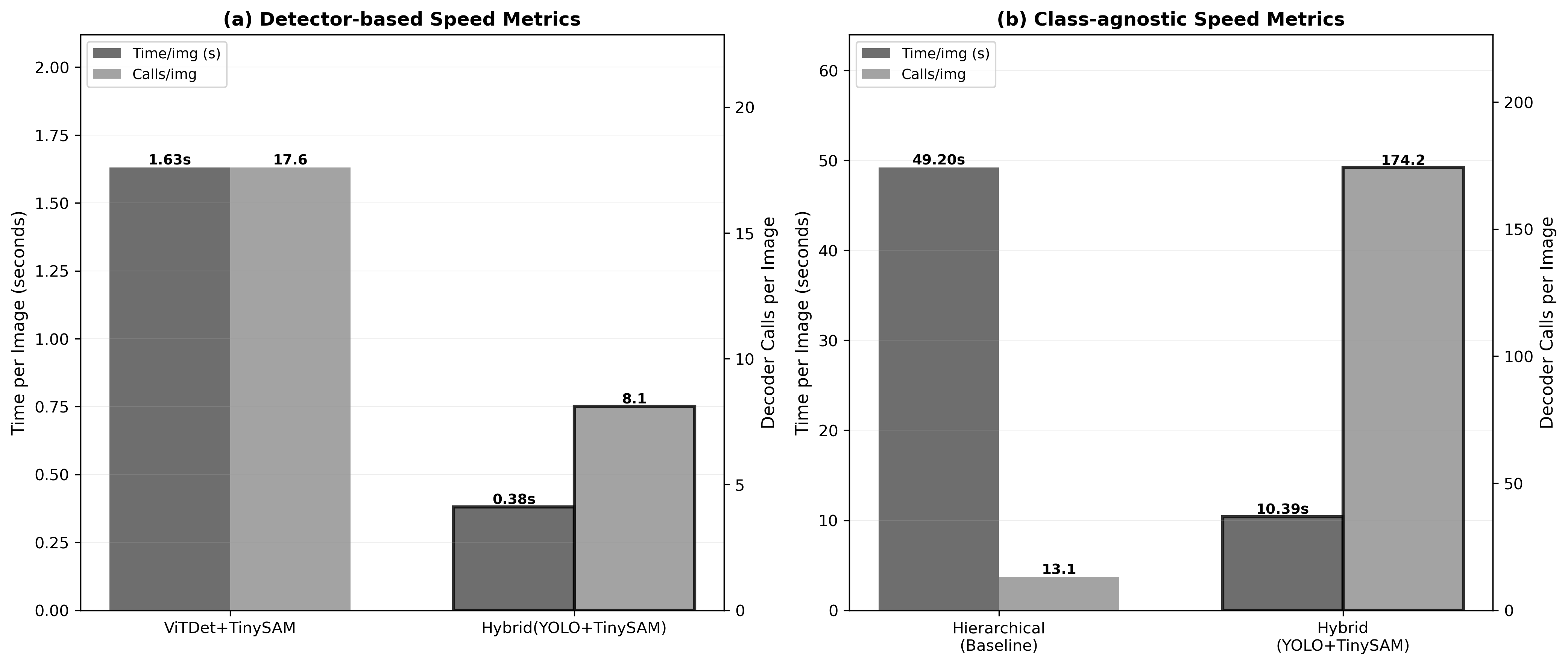}
\caption{Speed comparison.}
\label{fig:speed_comparison}
\end{figure}

\end{document}